# Path-based knowledge reasoning with textual semantic information for medical knowledge graph completion

Yinyu Lan[1,2], Shizhu He[1], Kang Liu[1], Xiangrong Zeng[3], Shengping Liu[3], Jun Zhao[1,2§]

[1] National Laboratory of Pattern Recognition, Institute of Automation, Chinese Academy of Sciences, Beijing, China

[2] School of Artificial Intelligence, University of Chinese Academy of Sciences, Beijing, China

[3] Beijing Unisound Information Technology Co., Ltd, Beijing, China

[§]Corresponding author

Email addresses:

    Yinyu Lan: yinyu.lan@nlpr.ia.ac.cn

    Shizhu He: shizhu.he@ nlpr.ia.ac.cn

    Kang Liu: kliu@ nlpr.ia.ac.cn

    Xiangrong Zeng: zengxiangrong@unisound.com

    Shengping Liu: liushengping@unisound.com

    Jun Zhao: jzhao@ nlpr.ia.ac.cn

# Path-based knowledge reasoning with textual semantic information for medical knowledge graph completion

## Abstract

**Background**

Knowledge graphs (KGs), especially medical knowledge graphs, are often significantly incomplete, so it necessitating a demand for medical knowledge graph completion (MedKGC). MedKGC can find new facts based on the exited knowledge in the KGs. The path-based knowledge reasoning algorithm is one of the most important approaches to this task. This type of method has received great attention in recent years because of high performance and interpretability. In fact, traditional methods such as path ranking algorithm (PRA) take the paths between an entity pair as atomic features. However, the medical KGs are very sparse, which makes it difficult to model effective semantic representation for extremely sparse path feature. The sparsity in the medical KGs is mainly reflected in the long-tailed distribution of entities and paths. Previous methods merely consider the context structure in the paths of knowledge graph and ignore the textual semantics of the symbols in the path. Therefore, their performance cannot be further improved due to the two aspects of entity sparseness and path sparseness.

**Methods**

To address the above issues, this paper proposes two novel path-based reasoning methods to solve the sparsity issues of entity and path respectively, which adopts the textual semantic information of entities and paths for MedKGC. By using the pre-trained model BERT, combining the textual semantic representations of the entities and the relationships, we model the task of symbolic reasoning in the medical KG as a numerical computing issue in textual semantic representation.

**Results**

Experiments results on the publicly authoritative Chinese symptom knowledge graph (CSKG) demonstrated that the proposed method is significantly better than the state-of-the-art path-based knowledge graph reasoning method, and the average performance is improved by 5.83% for all relations.

**Conclusions**

In this paper, we propose two new knowledge graph reasoning algorithms, which adopt textual semantic information of entities and paths and can effectively alleviate the sparsity problem of entities and paths in the MedKGC. As far as we know, it is the first method to use pre-trained language models and text path representations for medical knowledge reasoning. Our method can complete the impaired symptom knowledge graph in an interpretable way, and it outperforms the state-of-the-art path-based reasoning methods.

# Background

With the advent of the medical big data era, knowledge interconnection has received extensive attention [1]. How to extract useful medical knowledge from massive amounts of data is the key to medical big data analysis. Knowledge graph (KG) related technology provide one way to extract structured knowledge from massive texts and images. The combination of knowledge graph, big data and deep learning technology is the core driving force for the development of artificial intelligence. KG technology has also broad application prospects in the medical field [2], such as medical knowledge retrieval, auxiliary diagnosis and treatment, electronic medical records, etc. The application research of this technology in the medical field will play an important role in solving the contradiction between the insufficient supply of medical resources and the continuous increase in demand for medical services. KG is a graph that takes entities as nodes, and relations between entities as labeled-edges, which is usually stored in the form of inter-connecter triples (also called facts, one triple usually represent as (head entity, relation, tail entity)).

However, the widespread incompleteness of the KG greatly limited the effect of its application [3], and the downstream tasks cannot to be effectively supported due to lack of a large amount of facts. For this reason, a large number of knowledge graph completion (KGC) technologies have been proposed, which are trying to learn the reasoning model and infer new facts through the existing fact triples, and are an important task to solve the problem of incompleteness of knowledge graphs. At present, knowledge reasoning methods mainly include the following three categories: 1) Embedding methods translate entities and relations into a low-dimensional space, such as TransE [4], RESCAL [5], ComplEx [6], ANALOGY [7]. They achieve good results, but they only focus on the direct relations between entities and neglect the presence of indirect path relations in graphs; 2) Knowledge reasoning is a statistical relationship learning model that combines the probability graph model with the first-order predicate logic, such as Markov logic network and its variants [8, 9, 10]. Its core idea is bind weights to rules, which is able to soften the rigid constraints in the first-order predicate logic; 3) Path-based knowledge reasoning is a classifier model that learns the target relationship by taking the paths of of entities as features, such as, PRA [11], Path-RNN [12], Single-Model [13], Att-Model [14], etc.

Path-based knowledge reasoning methods have the advantages of good performance and interpretability, and at the same time, there is no need to add additional logic rules. This article mainly focuses on the research of such methods and is committed to improving the current path-based knowledge reasoning performance. In the knowledge graph, multiple triples can be connected through intermediate entities, and a path is usually defined as a sequence of entities and relationships. For example, as shown in **Figure 1**, <肺静脉畸形引流(anomalous pulmonaryvenous drainage), 疾病相关症状(disease-related symptoms), 呼吸窘迫(respiratory distress)> and <呼吸窘迫(respiratory distress), 症状相关科室 (symptom-related departments), 呼吸内科 (respiratory medicine)> form a path through the associated intermediate node "呼吸窘迫(respiratory distress)" . Based on this path, it can be inferred that there is a "疾病相关科室(disease-related departments)" relationship between "肺静脉畸形引流 (anomalous pulmonaryvenous drainage)" and "呼吸内科(respiratory medicine)" with paths such as "肺静脉畸形引流(anomalous pulmonaryvenous drainage)→疾病相关症状(disease-related symptoms)→鼓槌指(clubbing digits)→症状相关症状(symptom-related symptoms)→肺淋巴管肌瘤(pulmonary lymphangiomyomatosis)→症状相关科室 (symptom-related departments)→呼吸内科(respiratory medicine)" and "肺静脉畸形引流 (anomalous pulmonaryvenous drainage)→疾病相关症状(disease-related symptoms)→呼吸窘迫(respiratory distress)→症状相关疾病(symptom-related disease)→血气胸(hemopneumothorax)→疾病相关科室(disease-related departments)→呼吸内科(respiratory medicine)".

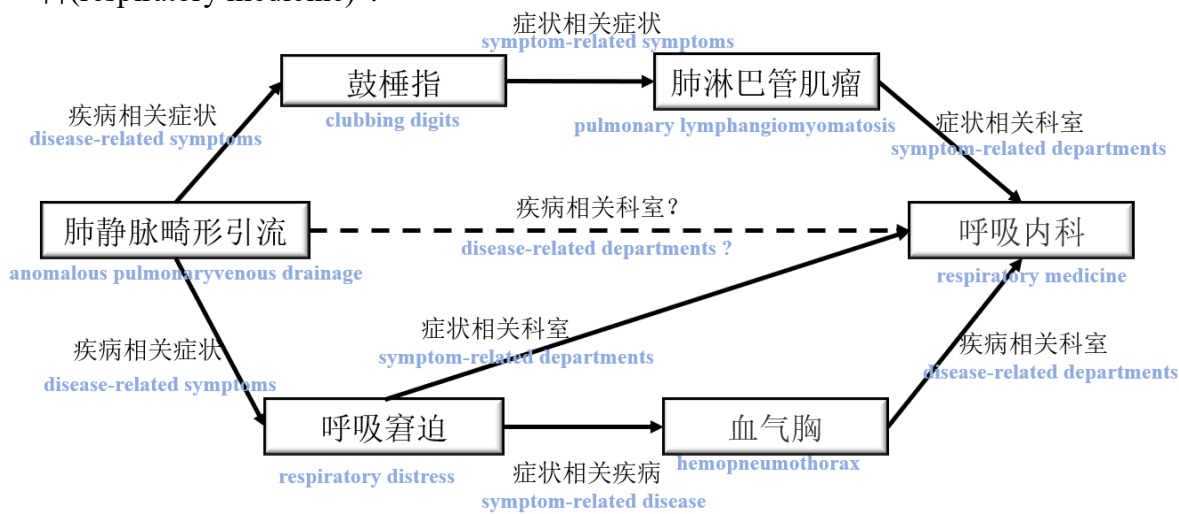

**Figure 1** A subgraph in the Chinese symptom knowledge graph. The rectangles represent entities, the solid edges between entities represent the relationship between the entities connected in the path, and the dotted edges represent the relationship that combines the information on multiple paths to determine whether there is a relationship.

However, the typical methods have some shortcomings. First of all, the previous method uses each path as an atomic feature [11], which results in a very large feature space that is difficult to train effectively. Secondly, previous methods take the paths as independent features and ignore their relationships of different atomic features. It can be seen from the **Figure 1**, that inferring relationships often need to rely on multiple paths between an entity pair, and different relations may have similar semantics, such as "症状相关科室(symptom-related departments)" and "疾病相关科室(disease-related departments)". Thirdly, previous methods only consider t the structural information for reasoning [12], without using the textual semantic information of the symbols. Even different paths may have similar semantics, for example, "肺静脉畸形引流(anomalous pulmonaryvenous drainage)→疾病相关症状(disease-related symptoms)→呼吸窘迫(respiratory distress)→症状相关疾病(symptom-related disease)→血气胸(hemopneumothorax)→疾病相关科室(disease-related departments)→呼吸内科(respiratory medicine)" and "肺静脉畸形引流(anomalous pulmonaryvenous drainage)→疾病相关症状(disease-related symptoms)→呼吸窘迫(respiratory distress)→症状相关科室(symptom-related departments)→呼吸内科(respiratory medicine)" own very close semantics.

Affected by the sparsity of the knowledge graph, it hinders the further improvement of the performance of traditional methods [3]. As shown in Figure 2, the paths and entities in the knowledge graph are very sparse and are distributed with long tails, and 35.56% of entities and 41.84% of paths only appeared once. Some recent studies [13, 14] began to combine multiple paths and began to incorporate entity information to enrich knowledge representation. However, they only considered the type information of the entity, in fact and the type of the entity is also different in different contexts. On the other hand, the textual information of entities and relationship also has rich semantic features, and it does not make full use of the syntax, grammatical patterns and semantic features of large-scale text data, so the performance cannot be further improved due to the two aspects of entity sparseness and path sparseness. The entities and relationships in the medical KG usually have names and labels in natural language, which can be combined into sentences. Therefore, an effective method to alleviate the above-mentioned sparsity problem is to use the textual semantic features of entities and relationships.

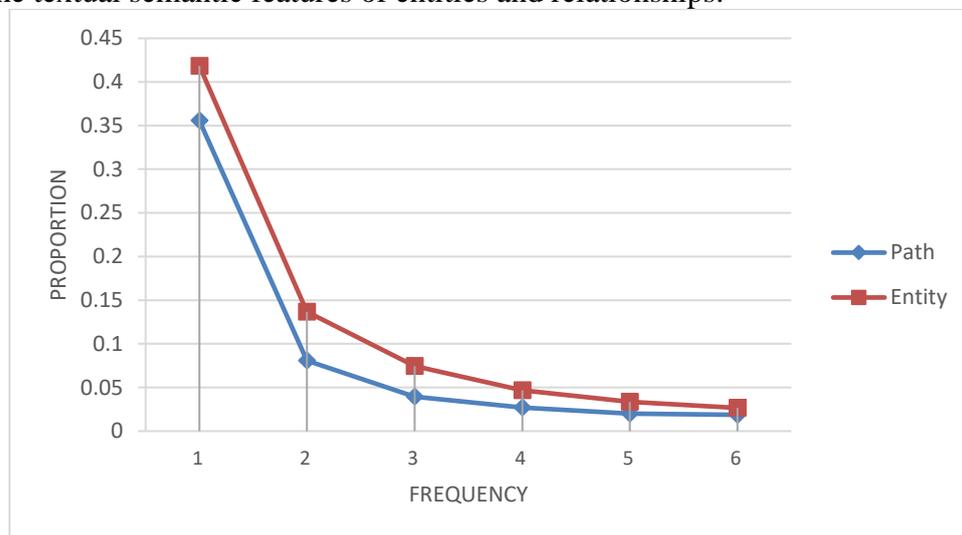

**Figure 2** Long-Tailed Distribution of Entities and Paths in Chinese Symptom Knowledge Graph

In fact, in the past two years, with the introduction of pre-trained language models such as ELMo [15], BERT [16], RoBERTa [17], XLNET [18], and GPT-3 [19], the semantic representation capabilities in general natural language processing tasks have made great progress. These models can learn high-quality contextual representation of words and sentences from a large amount of unstructured text data, and achieve state-of-the-art performance in many natural language understanding tasks. Among them, the most representative one is BERT, which uses the bidirectional transformer encoder for pre-training through masked language modeling (MLM) and next sentence prediction (NSP) tasks, which can capture rich semantic information in model parameters. For any natural language, pre-trained models such as BERT can supply numerical semantic representations with good generalization performance.

Therefore, in order to solve the shortcomings of traditional path-based knowledge reasoning methods and make full use of the semantic representation capabilities of pre-trained language models, this paper proposes two new knowledge graph reasoning algorithms based on the textual semantic representation of paths. Given an entity pair, and a set of paths between the entity pairs, we model the task of symbolic reasoning in the medical KG as a numerical computing issue in textual semantic representation, and using BERT encoding the statements of paths and entities text for capturing semantic features. We utilize the attention mechanism to learn the combined representation of multiple features, and then use the classifier model to predict whether there is a certain relationship between the entity pairs. The experimental results demonstrated that our method is 10.74% higher than the traditional PRA method on the public medical KG, and 5.83% higher than the previous best method.

## Methods

In this section, we first introduce language model pre-training and the overall framework of our models, and then introduces the details of the algorithm. Some symbols we may use in the algorithms: the entity pair to be queried is $(e_s, e_t)$, $\delta$ represents the query relationship, and the bold symbols denote the corresponding vector or matrix. $P_{(e_s, e_t)} = \{\pi_1, \pi_2, \pi_3 \ldots \pi_m\}$ represents the collection of paths between the entity pair $(e_s, e_t)$, $\pi = \{w_1, w_2, w_3 \ldots w_l\}$ represents a sequence of path textual statements, which is composed of the names and descriptions of the relationships and entities contained in the path.

**Language model pre-training**

The standard language model is to input a natural language text sequence by $W = [w_1, w_1, \ldots, w_n]$, and then output a probability about this sequence. Different from the traditional feature-based language model [26, 27], fine tuning approaches used the pre-trained model architecture and parameters as a starting point for specific NLP tasks. The Pre-trained models capture rich semantic patterns from free text, and achieve the best performance in many downstream tasks. Recently, pre-trained language models have also been explored in the context of KG. Wang et al. [20] learned the contextual embeddings on entity-relation chains (sentences) generated from random walks in KG, then used the embeddings as initialization of KG embeddings models like TransE [4]. Zhang et al. [21] incorporated informative entities in KG to enhance BERT language representation. By adding the names and descriptions of entities and relationships as input, Yao et al. [22] directly fine-tune BERT to calculate plausibility scores of triples without using the rich path information in the knowledge graph.

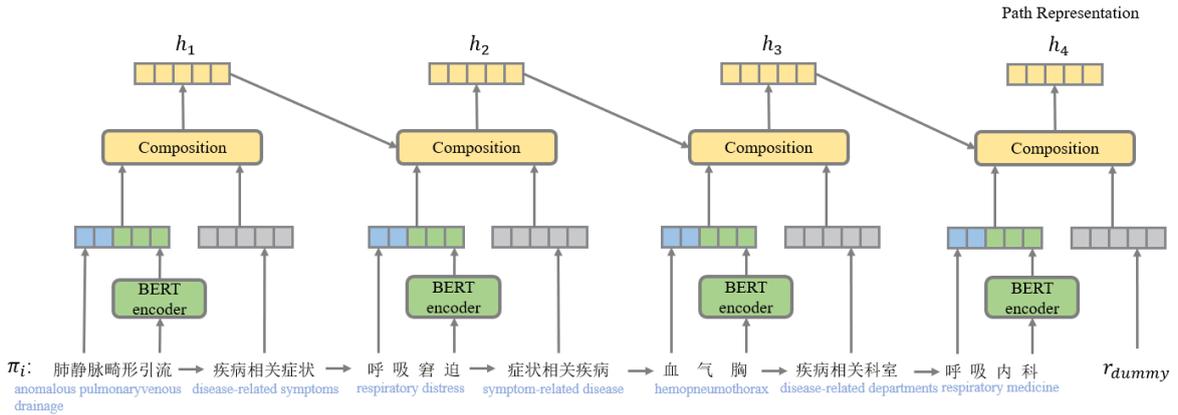

**Figure 3** The architecture of BERT enhanced entity representation used to extract path vector representation. $r_{dummy}$ is a dummy relation.

**Overall model framework**

On the basis of research [13, 14], this paper proposes the overall framework shown in **Figure 3** and **Figure 4**. Recently, there has also been research on how to represent knowledge as nature language [23, 24, 25]. On this basis, we use templates to represent entities and paths in CSKG into textual statement, for example, the entity textual statement of entity "枣树皮" (Jujube Bark) is "枣树皮，药品，中药。" (Jujube Bark, drug, traditional Chinese medicine.), and the path "肺静脉畸形引流 (anomalous pulmonaryvenous drainage)→疾病相关症状(disease-related symptoms)→呼吸窘迫 (respiratory distress)→症状相关科室(symptom-related departments)→呼吸内科 (respiratory medicine)" can be represented as "肺静脉畸形引流疾病的相关症状是呼吸窘迫，呼吸窘迫症状的相关科室是呼吸内科。(The related symptom of anomalous pulmonaryvenous drainage is respiratory distress, and the related department of respiratory distress is respiratory medicine.)". To make full use of the contextual representation with rich semantic information, we use BERT to encode entity textual statements and path textual statements to enhance the embedding of entities and paths. The path sequence is represented by each path sequence after the BERT encoding. The attention mechanism is used to combine the semantic features of multiple paths. The semantic similarity score between paths and query relation is finally used to determine whether there is a query relationship between entity pairs.

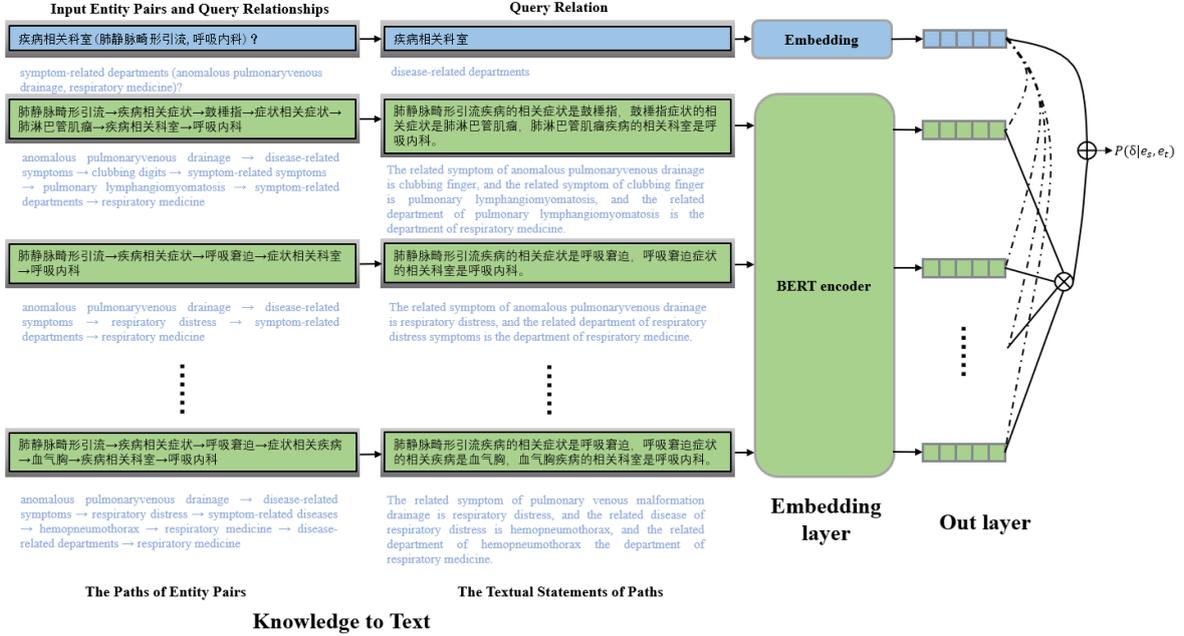

**Figure 4** The architecture of BERT enhanced path representation. The operation $\oplus$ denotes element-wise summation, and the operation $\otimes$ denotes weighted summation. The dotted line represents the attention mechanism.

**BERT enhanced entity representation**

As shown in **Figure 3**, in this module, each relation and entity in path is first mapped to a vector representation, and the entity type textual statement will be encoded, and their token representations are fed into the BERT model architecture, which is a multi-layer bidirectional transformer encoder based on the original implementation described in [28], to get the entity text representation. Then concatenated with the entity types embedding. The final hidden vector of the special [CLS] token is denoted as $\mathbf{C} \in \mathbb{R}^H$, where H is the hidden state size in pre-trained BERT. The final hidden state $\mathbf{C}$ corresponding to [CLS] is used as the entity statement representation:

$$\mathbf{C}_{t-1} = BERT(\mathrm{ed}_{t-1}) \qquad (1)$$

$$\hat{e}_{t-1} = e_{t-1} \bigcup \mathbf{C}_{t-1} \qquad (2)$$

Where $\mathrm{ed}_{t-1}$ denotes entity textual statement, and the operation $\bigcup$ denotes concatenating two vectors, $\mathbf{C}_{t-1} \in \mathbb{R}^H$.

Then entity representation and relationship representation are composed sequentially in an RNN. At each RNN step t, the model consumes the representation of entity $\mathbf{e}_{t-1}$ ($\mathbf{e}_0 = \mathbf{e}_s$) and a relation $\mathbf{r}_t$, and outputs a hidden state $\mathbf{h}_t$. To resist the sparseness of the entity and reduce model parameters, we map each entity to the averaged representation of its types. For simplicity, we still use $\mathbf{e}_{t-1} \in \mathbb{R}^{d \times d}$ to denote the averaged type representation of entity $\mathrm{e}_{t-1}$. Here $\mathbf{r}_t \in \mathbb{R}^d$, $\mathbf{h}_t \in \mathbb{R}^d$, the RNN hidden state is given by:

$$\mathbf{h}_t = f(\mathbf{W}_1 \mathbf{h}_{t-1} + \mathbf{W}_2 r_{t-1} + \mathbf{W}_3 \hat{\mathbf{e}}_{t-1}) \qquad (3)$$

where $\mathbf{W}_1 \in \mathbb{R}^{d \times d}$, $\mathbf{W}_2 \in \mathbb{R}^{d \times d}$, $\mathbf{W}_3 \in \mathbb{R}^{d \times k}$ are RNN parameter matrices. $f$ is a non-linear function. In the proposed method, $f = ReLU()$. as shown in Equation (4), The context representation of entity pairs is given by:

$$\mathbf{ep}_{s,t}^{\delta} = f\left(\sum_{n=1}^{N} \alpha_i^{\delta} \boldsymbol{\pi}_i\right) \quad (4)$$

where $\alpha_i^{\delta}$ is the weight of path i when modelling the entity pair representation for query relation δ, and $f = Tanh()$. The weight for each path is as follow:

$$\alpha_i^{\delta} = \frac{exp(z_i^{\delta})}{\sum_j exp(z_j^{\delta})} \quad (5)$$

Where $z_j^{\delta}$ measures how well input path $\pi_i$ and query relation δ matches, and is as follow:

$$z_j^{\delta} = f(\boldsymbol{\pi}_i \mathbf{T})\boldsymbol{\delta} \quad (6)$$

Where $f = Tanh()$, $\mathbf{T} \in \mathbb{R}^{d \times d}$. After getting the query statement representation and the path context representation of the entity pair, calculate the probability that the entity pair has the query relationship:

$$P(\delta|e_s, e_t) = \sigma(ep_{s,t} \cdot \delta) \quad (7)$$

Where σ is sigmoid function. Following Das et al. [14], we train a single model for all query relations. The model is trained to minimize the negative log-likelihood, and the simplified form of the objective function is defined as follows:

$$L\left(\Theta, \Delta_R^+, \Delta_R^-\right) = -\sum_{e_s,e_t,\delta \in \Delta_R^+} logP(\delta|e_s, e_t) - \sum_{\widehat{e_s},\widehat{e_t},\widehat{\delta} \in \Delta_R^-} log\left(1 - P(\widehat{\delta}|\widehat{e_s}, \widehat{e_t})\right) \quad (8)$$

Where $\Delta_R^+$ denotes the set of positive triples and $\Delta_R^-$ denotes the set of negative triples. We also use the standard L2 norm of weights as a constraint function. The model parameters are randomly initialized and updated by considering a gradient step with a constant learning rate on the batch of training triples. In our experiment, we apply a range of learning rates to find out how this affects prediction performance. The training is stopped when the loss function converges to an optimal point.

**BERT enhanced path representation**

Take each sentence sequence π in the path set of the entity pair. The first position of the sequence is inserted by the classification mark symbol [CLS], and the last position is inserted by the [SEP] symbol to represent the end of the sequence. After the BERT encoding, taking the output final hidden layer representation of [CLS] symbolic as the embedding of the path sequence, we can get the set of path textual statement representation $P_{(e_s,e_t)} = \{\boldsymbol{\pi}_1, \boldsymbol{\pi}_2, ..., \boldsymbol{\pi}_n\}$, $\boldsymbol{\pi} \in \mathbb{R}^d$. For example, The input path text is "[CLS]肺静脉畸形引流疾病的相关症状是呼吸窘迫，呼吸窘迫症状的相关科室是呼吸内科。[SEP]" (The related symptom of anomalous pulmonaryvenous drainage is respiratory distress, and the related department of respiratory distress symptoms is the department of respiratory medicine.), and it is fed into the BERT model as follow:

$$\pi_i = BERT(pd_i) \quad (9)$$

Where $pd_i$ is the input path text. We use the final hidden vector of [CLS] token to represent the path representation $\pi_i$. Then, like BERT enhanced entity representation, it uses the attention mechanism to combine multiple path information, and uses the same output layer and objective function.

## Experiments

In this section, we first introduce the dataset and the details of experiment data preparation, followed by the metric (mean average precision, MAP) used to measure the performance of our methods and the baseline methods for relation classification. Then, hyperparameter settings and overall experimental results as well as comparison results in each relationship are introduced. Finally, we present several cases to embody the effectiveness of the attention mechanism and the interpretability of reasoning.

**Dataset**

OpenKG is an open source knowledge graph community project advocated by Chinese Information Processing Society of China, it provides a large number of open source knowledge graph resources. The Chinese symptom knowledge graph in the OpenKG was the main resource for our work, and we obtain the path by random walks (RWs) to construct the experimental dataset, which we named CSKG.

**Data preparation**

This article builds an experimental dataset on the public Chinese symptom knowledge graph, and uses the random walk method to obtain the path between entity pairs. For negative examples, we randomly replacing the head entity, tail entity, and relationship in the triple with a uniformly sampled random entity or relation. In order to test and evaluate the ability of our proposed model to distinguish negative examples with the same relationship, which greatly increases the difficulty of the model to distinguish between positive and negative examples, when we randomly destroy entities, 70% probability to choose entities with the same relationship as query relation. Models in comparison are all evaluated on a subset of facts hidden during training. Training set, validation set, test set are separated randomly according to the ratio of 7:1.5:1.5. In this dataset, the number of paths between an entity pair ranges drastically from 1 to 622, so the robust of methods in comparison can be better evaluated with this dataset. Statistics of CSKG dataset is listed in **Table 1**.

**Table 1** Statistics of CSKG dataset

| Stats | Number |
| --- | --- |
| # CSKG triples | 629,538 |
| # Relation types | 17 |
| # Entities | 59,881 |
| # Paths | 28M |
| Avg. paths/query relation | 1.68M |
| Avg. path length | 3.88 |
| Max path length | 7 |
| Avg. training positive instances/query relation | 19,799 |
| Avg. training negative instances/query relation | 14,929 |
| Avg. positive test instances/query relation | 4,242 |
| Avg. negative test instances/query relation | 34,211 |

**Comparative experiment with baseline models**
- **PRA** [11]: This was the first method to implement path-based reasoning. It was presented by Lao et al. [11]. It uses distinct features to represent the paths that connect entities, creates a large feature matrix, and then trains a binary classification model on the feature matrix.
- **Path-RNN** [12]: is a model using RNN to predict binary target relations on the collected path sequences.
- **Single-Model** [13]: is an improved RNN model based on Path-RNN, which considers one model for all query relations, and utilizes LogSumExp, which is a smooth

- 9 -

approximation to max operation, to conduct score pooling for multiple paths.
- **Single-Model + Types** [13]: is the best model achieved by Das et al. [13], which represents entities as a combination of entities and an average function of all the entity types.
- **Att-model** [14]: is a model that using attention mechanism instead of LogSumExp for multiple paths between entity pairs compared with single model.
- **Att-Model + Types** [14]: is an improved model based on Att-Model with entities represented as a combination of entities and an average function of all the entity types.

**Evaluation metrics**

We use MAP as evaluation metrics, following recent works [13, 14] evaluating knowledge graph completion performance. MAP is the average of precision values at the ranks where relevant correct entities are ranked. The MAP score is computed using the following equation:

$$\text{MAP} = \frac{1}{|Q_r|} \sum_{q \in Q_r} \text{AP}(q) \tag{9}$$

Where $Q_r$ is the set of relationship types, AP is the average of precision scores at the rank locations of each correct result.

**Implementation details**

We set the baseline model according to the best performance configuration in the original paper. All model parameters to be learned are initialized randomly, and the optimization method is Adam. Hyper-parameters of each model are tuned on development set, and training is stopped when the accuracy on the development set does not improve by 0.01 within the last 10 epochs. We apply a grid search approach to tune the hyperparameters in our model. We select the learning rate, $\gamma$, for the Adam optimizer among {0.0001, 0.001, 0.002, 0.0025, 0.003}, the dimension of relation representation and the hidden states $d$, $h$ among {50, 100, 150, 200, 250, 300}, and the dimension of entity type $m$ among {50, 100, 100, 150, 200, 250, 300}. Model are trained for 100 epochs, with batch size = 64, learning rate = $1e^{-3}$, and l2-regularizer $\lambda = 1e^{-5}$. Adam settings are as default: $\beta_1 = 0.9$; $\beta_2 = 0.999$; $\epsilon = 1e^{-8}$.

**Experimental results**

We test the effectiveness of our method on 17 query relations, and report the results in **Table 2**.

Table 2 Experiments results on CSKG dataset

| Model | %MAP |
|---|---|
| PRA | 43.78 |
| Path-RNN | 43.83 |
| Single-Model | 45.93 |
| Att-Model | 46.37 |
| Single-Model + Types | 48.24 |
| Att-Model + Types | 48.85 |
| BERT enhanced entity representation | 51.90 |
| BERT enhanced path representation | **54.52** |

From the results, we can observe that our algorithm achieves the best performance. Specifically, (1) The experiment of BERT enhanced path representation demonstrates the superiority of our methods compared to other models after fusing the textual semantics of all entities and relationships. Our method achieves the best results, which is 5.83% higher than the previous best method, Att-Model + Types, which demonstrates that the inference performance can indeed be further improved after adopting textual semantic information of paths, which effectively alleviate the sparsity problem of paths; (2) Bert enhanced path representation is also 2.05% higher than the previous best method. It shows that only incorporating the textual semantics of entity types can also alleviate the problem of entity sparsity. PRA and Path-RNN suffers significantly, because of that treats each query relation separately. Single-Model and Att-model suffers from the sparseness of KG, and cannot surpass our methods.

To better show the strength and weakness of the proposed methods against Single-Model + Types and Att-model + Types, we further make a more detailed comparison for each relation. First, we compare the MAP scores of several methods on 17 relationships in the dataset. The results are listed in **Table 3**. It can be observed that with our methods achieved the best performance in all relationships. Among them, the "疾病相关部位 (disease-related body parts)" category has the largest improvement 22.42% (from 44.56% to 65.55%). This result fully demonstrates that our methods improves the shortcomings of Single-Model and Att-model.

**Table 3** %MAP performance on each relation

| Relations | Single-Model + Types | Att-Model + Types | BERT enhanced entity representation | BERT enhanced path representation |
|---|---|---|---|---|
| 检查相关症状 (Examination-related symptoms) | 38.58 | 50.79 | **53.31**(+2.52) | 38.84(-11.95) |
| 检查相关部位 (Examination-related body parts) | 52.90 | 51.64 | 52.75(-0.15) | **70.01**(+17.11) |
| 疾病相关症状 (Disease-related symptoms) | 34.35 | 45.51 | **48.43**(+2.92) | 38.56(-6.95) |
| 疾病相关科室 (Disease-related departments) | 43.56 | 42.27 | 47.23(+3.67) | **56.24**(+12.68) |
| 检查相关检查 (Examination-related examinations) | 57.35 | 49.85 | 53.88(-3.47) | **58.29**(+0.94) |
| 症状相关疾病 (Symptom-related diseases) | 39.54 | 46.81 | 47.12(+0.31) | **49.48**(+2.67) |
| 疾病相关检查 (Disease-related examinations) | 43.64 | 38.28 | 38.23(-5.41) | **56.42**(+12.79) |
| 症状相关科室 (Symptom-related departments) | 50.55 | 51.49 | **57.24**(+5.79) | 53.67(+2.18) |
| 症状相关症状 (Symptom-related symptoms) | 57.59 | 71.44 | **73.78**(+2.34) | 48.75(-22.69) |
| 疾病相关疾病 (Disease-related diseases) | 37.00 | 44.33 | **48.61**(-4.28) | 34.07(-10.26) |
| 疾病相关药品 (Disease-related drugs) | 51.29 | 47.07 | 56.34(-5.05) | **58.61**(+7.32) |
| 症状相关部位 (Symptom-related body parts) | 42.55 | 39.86 | 42.34(-0.21) | **47.45**(+4.9) |
| 检查相关科室 (Examination-related departments) | 44.56 | 40.16 | 41.43(-3,13) | **65.55**(+20.99) |
| 症状相关检查 (Symptom-related departments) | 53.50 | 47.53 | 56.73(+3.23) | **65.34**(+11.84) |
| 检查相关疾病 (Examination-related diseases) | 58.17 | 56.05 | **58.43**(+0.26) | 43.80(-14.37) |
| 疾病相关部位 (Disease-related body parts) | 58.58 | 56.95 | 57.66(-0.92) | **81.00**(+22.42) |
| 症状相关药品 (Symptom-related drugs) | 56.28 | 50.39 | 48.81(-7.39) | **63.40**(+7.12) |

Note: Bold font shows best performance achieved in the experimental models. The value in parentheses indicates the percentage increase compared to the best score between Single-Model + Types and Att-Model + Types.

**Case study**

In this section, we use two cases to embody the effectiveness of using the attention mechanism and the interpretability of reasoning. We choose the query "症状相关症状(两眼上视障碍, 耳聋)?" (Symptom-related symptoms(Binocular superior visual impairment, Epicophosis)?) and "疾病相关药品(糖尿病所致骨髓病, 甲酚皂溶液)?" (Disease-related diseases(Bone marrow disease caused by diabetes, Cresol soap solution)?),and select two of the positive examples. Then we observe the attention weights separately. High attention weight and low attention weight case for path textual statement are shown in **Table 4**. It can be seen from the table, that the weight of the path textual statement closer to the query semantics will be higher, while the path textual statement with low attention tend to lack the ability of prediction.

**Table 4** Examples of attention mechanism in CSKG dataset

| | |
|---|---|
| Query | 症状相关症状(两眼上视障碍，耳聋)?<br>Symptom-related symptoms(Binocular superior visual impairment, Epicophosis)? |
| High weight | 两眼上视障碍症状的相关症状是听觉下降，听觉下降症状的相关症状是耳聋。<br>The related symptom of the symptoms of upper binocular vision disorder is hearing loss, the related symptom of hearing loss is deafness. |
| Low weight | 两眼上视障碍症状的相关疾病是偏头风，偏头风疾病的相关疾病是小儿偏头痛，小儿偏头痛疾病的相关症状是复视，复视症状的相关症状是耳聋。<br>The related disease of the symptoms of visual disturbance in both eyes is migraine, the related disease of migraine is migraine in children, and the related symptom of migraine in children is diplopia, and the related symptom of diplopia is deafness. |
| Query | 疾病相关药品(糖尿病所致骨髓病，甲酚皂溶液)?<br>Disease-related diseases(Bone marrow disease caused by diabetes, Cresol soap solution)? |
| High weight | 糖尿病所致骨髓疾病的相关症状是脊髓病变，脊髓病变的相关药品是甲酚皂溶液。<br>The related symptom of bone marrow disease caused by diabetes is spinal cord lesions, and the related medicine for spinal cord lesions is cresol soap solution. |
| Low weight | 糖尿病所致骨髓疾病的相关疾病是周围神经病损，周围神经病损疾病的相关症状是感觉过敏，感觉过敏症状的相关疾病是神劳，神劳疾病的相关症状是无力，无力症状的相关疾病是重症肌无力危象，重症肌无力疾病相关药品是甲酚皂溶液。<br>The related disease of bone marrow disease caused by diabetes is peripheral neuropathy, the related symptom of peripheral neuropathy is hyperesthesia, the related disease of hyperesthesia is mental fatigue, the related symptom of mental fatigue is weakness, and the related disease of weakness is myasthenia gravis, and the related medicine for myasthenia crisis is cresol soap solution. |

## Discussion

Experimental results have demonstrated the superiority of our model in both reasoning effectiveness and interpretability, which is the first attempt to employ BERT and textual path representations for MedKGC. There is a limitation affecting our works. The huge number of parameters of BERT will reduce the speed of model training and inference. But we think this is a trade-off for better performance. By applying knowledge distillation [29] technology, this problem can be alleviated, and we leave this for future research. In the future work, we will consider further exploring the joint knowledge graph structure and text information for modeling, which is a direction worth studying. At the same time, we will focus on language models pre-training with more text data, such as GPT-3. In addition, we are also preparing to apply our methods to more tasks related to medical knowledge graph reasoning, such as medical knowledge graph question answering.

## Conclusions

This paper points out the shortcomings of current path-based reasoning methods, and proposes two new medical knowledge graph reasoning algorithms based on the textual semantic representation of paths, which effectively alleviate the problem that the sparseness of entities and paths in the medical KG. In our experiments, we show that our method performs better than recent state-of-the-art methods on MedKGC task and can efficiently represent the paths between an entity pair to predict their missing relation. We use the pre-trained language model to enhance entities and paths representations, and the attention mechanism is used to combine the semantic features of multiple paths. We conducted an empirical evaluation of this method over a public challenging medical KG, and the experimental results have demonstrated that our method has better performance than previous path-based relational reasoning methods. We believe that integrating text information of entities and relationships, by the large number of text semantic patterns encoded in the pre-trained language model, is a promising approach for medical knowledge reasoning.

## Abbreviations
**KG**: Knowledge Graph; **KGE**: Knowledge Graph Embedding; **KGC**: Knowledge Graph completion; **RW**: Random Walk; **CSKG**: Chinese symptom knowledge graph; **MLM**: Masked Language Modeling; **NSP**: Next Sentence Prediction; **PRA**: path ranking algorithm

# Declarations

## Authors' contributions
Author leaded the method application, experiment conduction and the result analysis. Author participated in the data extraction and preprocessing. Author participated in the manuscript revision. Author provided theoretical guidance and the revision of this paper.

## Ethics approval and consent to participate
Not applicable.

## Consent for publication
Not applicable.

## Availability of data and materials
The datasets used and analyzed during the current study are available from the first author upon reasonable requests. / The dataset is publically available via XXX.

## Competing interests
The authors declare that they have no competing interests.

## Funding
This work is supported by the Natural Key R&D Program of China (No.2017YFB1002101), the National Natural Science Foundation of China (No.61922085, No.61976211, No.61702512) and the Key Research Program of the Chinese Academy of Sciences (Grant NO. ZDBS-SSW-JSC006). This research work was also supported by the independent research project of National Laboratory of Pattern Recognition and the Youth Innovation Promotion Association CAS.

## References

1. Bello-Orgaz G, Jung JJ, Camacho D. Social big data: Recent achievements and new challenges. *Information Fusion*, 2016; 28:45-59.
2. Murdoch TB, Detsky AS. The inevitable application of big data to health care. *JAMA*. 2013; 309(13): 1351-1352.
3. Pujara J, Augustine E, Getoor L. Sparsity and noise: Where knowledge graph embeddings fall short. In: Proceedings of the 2017 Conference on Empirical Methods in Natural Language Processing; Association for Computational Linguistics. Copenhagen; 2017. p. 1751-1756.

4. Bordes A, Usunier N, Garcia-Duran A, Weston J, Yakhnenko O. Translating embeddings for modeling multi-relational data. Advances in neural information processing systems. Harrahs and Harveys, Lake Tahoe; 2013. p. 2787-2795.
5. Nickel M, Tresp V, Kriegel HP. A three-way model for collective learning on multi-relational data. In: International Conference on Machine Learning (ICML). Bellevue; 2011. p. 11, 809-816.
6. Trouillon T, Welbl J, Riedel S, Gaussier É, Bouchard G. Complex embeddings for simple link prediction. In: International Conference on Machine Learning (ICML). New York; 2016. p. 2071-2080.
7. Liu H, Wu Y, Yang Y. Analogical inference for multi-relational embeddings. arXiv preprint arXiv:1705.02426, 2017.
8. Chen DZWY, Wang DZ. Web-scale knowledge inference using markov logic networks. In: ICML workshop on Structured Learning: Inferring Graphs from Structured and Unstructured Inputs. Association for Computational Linguistics. Atlanta; 2013. p. 106-110.
9. Jiang S, Lowd D, Dou D. Learning to refine an automatically extracted knowledge base using markov logic. In: 2012 IEEE 12th International Conference on Data Mining. Berkeley: IEEE; 2012. p. 912-917.
10. Pujara J, Miao H, Getoor L, Cohen W. Knowledge graph identification. In: International Semantic Web Conference. Springer, Berlin, Heidelberg; 2013. p. 542-557.
11. Lao N, Mitchell T, Cohen W. Random walk inference and learning in a large scale knowledge base. In: Proceedings of the 2011 conference on empirical methods in natural language processing. Association for Computational Linguistics. Edinburgh; 2011. p. 529-539.
12. Neelakantan A, Roth B, McCallum A. Compositional vector space models for knowledge base completion. arXiv preprint arXiv:1504.06662, 2015.
13. Das R, Neelakantan A, Belanger D, et al. Chains of reasoning over entities, relations, and text using recurrent neural networks. arXiv preprint arXiv:1607.01426, 2016.
14. Jiang X, Wang Q, Qi B, Qiu Y, Li P, Wang B. Attentive path combination for knowledge graph completion. In: Asian conference on machine learning. Seoul; 2017. p. 590-605.
15. Peters ME, Neumann M, Iyyer M, Gardner M, Clark C, Lee K, Zettlemoyer L. Deep contextualized word representations. arXiv preprint arXiv:1802.05365, 2018.
16. Devlin J, Chang MW, Lee K, et al. Bert: Pre-training of deep bidirectional transformers for language understanding. arXiv preprint arXiv:1810.04805, 2018.
17. Liu Y, Ott M, Goyal N, Du J, Joshi M, Chen D, Stoyanov V. Roberta: A robustly optimized bert pretraining approach. arXiv preprint arXiv:1907.11692, 2019.
18. Yang Z, Dai Z, Yang Y, Carbonell J, Salakhutdinov RR, et al. Xlnet: Generalized autoregressive pretraining for language understanding. In: Advances in neural information processing systems. Vancouver Convention Center, Vancouver; 2019. p. 5753-5763.
19. Brown T B, Mann B, Ryder N, Subbiah M, Kaplan J, Dhariwal P, et al. Language models are few-shot learners. arXiv preprint arXiv:2005.14165, 2020.
20. Wang H, Kulkarni V, Wang WY. Dolores: Deep contextualized knowledge graph embeddings. arXiv preprint arXiv:1811.00147, 2018.
21. Zhang Z, Han X, Liu Z, Jiang X, Sun M, Liu Q. ERNIE: Enhanced language representation with informative entities. arXiv preprint arXiv:1905.07129, 2019.
22. Yao L, Mao C, Luo Y. KG-BERT: BERT for knowledge graph completion. arXiv preprint arXiv:1909.03193, 2019.

23. Chisholm A, Radford W, Hachey B. Learning to generate one-sentence biographies from wikidata. arXiv preprint arXiv:1702.06235, 2017.
24. Kale M. Text-to-Text Pre-Training for Data-to-Text Tasks. arXiv preprint arXiv:2005.10433, 2020.
25. Tylenda T, Kondreddi SK, Weikum G. Spotting knowledge base facts in web texts Proceedings of the 4th Workshop on Automated Knowledge Base Construction. Montreal; 2014. p. 1-6.
26. Mikolov T, Sutskever I, Chen K, Corrado GS, Dean J. Distributed representations of words and phrases and their compositionality. In: Advances in neural information processing systems. Harrahs and Harveys, Lake Tahoe; 2013. p. 3111-3119.
27. Pennington J, Socher R, Manning CD. Glove: Global vectors for word representation. In: Proceedings of the 2014 conference on empirical methods in natural language processing (EMNLP). Doha, Qatar; 2014. p. 1532-1543.
28. Vaswani A, Shazeer N, Parmar N, Uszkoreit J, Jones L, Gomez AN, et al. Attention is all you need. In: Advances in neural information processing systems. Long Beach Convention Center, Long Beach; 2017. p. 5998-6008.
29. Hinton G, Vinyals O, Dean J. Distilling the knowledge in a neural network. arXiv preprint arXiv:1503.02531, 2015.